\documentclass{article}
\usepackage{amsmath}
 \usepackage{multirow}

\usepackage{listings}
\usepackage{color} 

\definecolor{codegreen}{rgb}{0,0.6,0}
\definecolor{codegray}{rgb}{0.5,0.5,0.5}
\definecolor{codepurple}{rgb}{0.58,0,0.82}
\definecolor{backcolour}{rgb}{0.95,0.95,0.92}

\lstdefinestyle{mystyle}{
    backgroundcolor=\color{backcolour},   
    commentstyle=\color{codegreen},
    keywordstyle=\color{magenta},
    numberstyle=\tiny\color{codegray},
    stringstyle=\color{codepurple},
    basicstyle=\footnotesize,
    breakatwhitespace=false,         
    breaklines=true,                 
    captionpos=b,                    
    keepspaces=true,                 
    numbers=left,                    
    numbersep=5pt,                  
    showspaces=false,                
    showstringspaces=false,
    showtabs=false,                  
    tabsize=2
}

\usepackage[preprint]{neurips_2024}

\usepackage{natbib}
\usepackage[utf8]{inputenc} 
\usepackage[T1]{fontenc}    
\usepackage{hyperref}       
\usepackage{url}            
\usepackage{booktabs}       
\usepackage{amsfonts}       
\usepackage{nicefrac}       
\usepackage{microtype}      
\usepackage{xcolor}         
\usepackage{color}
\definecolor{pink}{RGB}{255,170,182}

\usepackage{todonotes}
\newcommand{\xiaoshuicomment}[1]{\todo[color=purple!20, inline, author=USER]{#1}}

\title{Accessing GPT-4 level Mathematical Olympiad Solutions via Monte Carlo Tree Self-refine with LLaMa-3 8B: A Technical Report}

\author{%
  Di~Zhang\\
Fudan University\\
  Shanghai Artificial Intelligence Laboratory\\
  \texttt{di.zhang@ustc.edu} \\
  \And
  Xiaoshui~Huang\\
Shanghai Artificial Intelligence Laboratory\\
  \texttt{xiaoshuihuang2019@gmail.com} \\
  \And
  Dongzhan~Zhou\\
Shanghai Artificial Intelligence Laboratory\\
  \texttt{zhoudongzhan@pjlab.org.cn} \\
    \And
  Yuqiang~Li\\
Shanghai Artificial Intelligence Laboratory\\
  \texttt{liyuqiang@pjlab.org.cn} \\
    \And
  Wanli~Ouyang\\
Shanghai Artificial Intelligence Laboratory\\
  \texttt{wanli.ouyang@sydney.edu.au} \\
}

\begin{document}

\maketitle

\begin{abstract}
This paper introduces the MCT Self-Refine (MCTSr) algorithm, an innovative integration of Large Language Models (LLMs) with Monte Carlo Tree Search (MCTS), designed to enhance performance in complex mathematical reasoning tasks. Addressing the challenges of accuracy and reliability in LLMs, particularly in strategic and mathematical reasoning, MCTSr leverages systematic exploration and heuristic self-refine mechanisms to improve decision-making frameworks within LLMs. The algorithm constructs a Monte Carlo search tree through iterative processes of Selection, self-refine, self-evaluation, and Backpropagation, utilizing an improved Upper Confidence Bound (UCB) formula to optimize the exploration-exploitation balance. Extensive experiments demonstrate MCTSr's efficacy in solving Olympiad-level mathematical problems, significantly improving success rates across multiple datasets, including GSM8K, GSM Hard, MATH, and Olympiad-level benchmarks, including Math Odyssey, AIME, and OlympiadBench. The study advances the application of LLMs in complex reasoning tasks and sets a foundation for future AI integration, enhancing decision-making accuracy and reliability in LLM-driven applications. Codes publicly accessible at \href{https://github.com/trotsky1997/MathBlackBox}{github.com/trotsky1997/MathBlackBox}.
\end{abstract}

\section{Introduction}
\begin{figure}[htbp]
    \centering
    \includegraphics[width=1\linewidth]{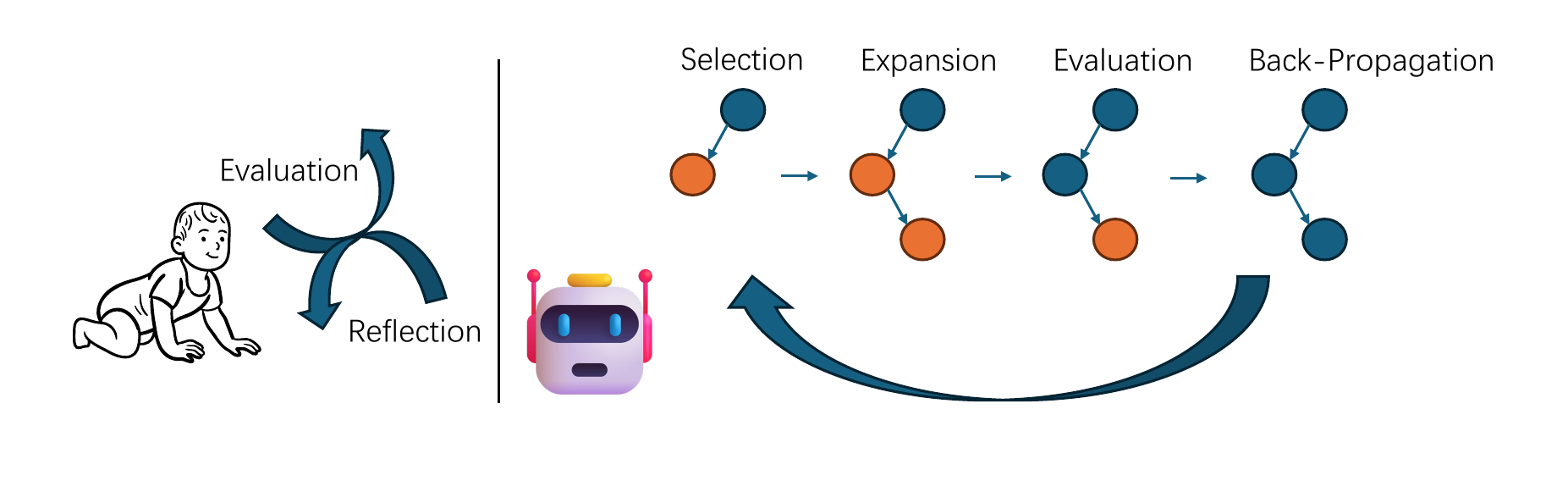}
    \caption{Agents can learn decision-making and reasoning from the trial-and-error as humans do.}
    \label{fig:main pic}
\end{figure}
 
With the rapid evolution of artificial intelligence, large language models (LLMs) such as GPT-4~\citep{achiam2023gpt} and LLaMA~\citep{touvron2023llama} have become fundamental in advancing natural language processing (NLP) capabilities. These models, characterized by their multi-billion parameter architectures, exhibit remarkable language comprehension and generation abilities. Their emergent properties, including reasoning and in-context learning, have opened new avenues for addressing complex NLP tasks beyond traditional domains, encompassing mathematical problem-solving~\citep{yu2023metamath,yuan2023scaling}, recommendation systems~\citep{Lyu2023LLMRecPR}, and even molecule generation~\citep{Liang2023DrugChatTE}. However, despite these advancements, LLMs face notable challenges in areas demanding strategic and logical reasoning.

One significant hurdle is the accuracy and trustworthiness of the outputs. Especially in mathematical contexts, where precision is paramount, the reasoning capabilities of LLM always suffer from prone to producing hallucinations—outputs~\citep{Huang2023ASO} that, while superficially plausible, but irrelevant or factually incorrect, are finally harmful to rational processes. Though rewriting techniques like Self-Refine~\citep{Madaan2023SelfRefineIR} can help relieve, this tendency can still lead to misleading or wrong outcomes in real-world complex mathematical problems. 

To address these challenges, this paper proposes MCT Self-Refine (MCTSr), an integration of LLMs with a Monte Carlo Tree Search (MCTS) algorithm~\citep{Chaslot2008MonteCarloTS}, focusing on enhancing LLMs' performance in complex mathematical reasoning tasks, such as those encountered in mathematical Olympiads. MCTS, a decision-making tool widely used in AI for scenarios requiring strategic planning, is typically employed in gaming and complex problem-solving environments. By combining MCTS's systematic exploration capabilities with LLMs' capabilities of Self-Refine and Self-Evaluation, we aim to create a more robust framework for tackling intricate reasoning tasks that current LLMs struggle with.

Several technical challenges exist in adapting MCTS for LLM integration. Traditional MCTS strategies may not align well with the stochastic and generative nature of LLM outputs, which often involve an infinite, continuous space of potential actions. This misalignment necessitates a tailored approach to expectation calculation and Backpropagation within the MCTS framework to better suit the unique characteristics of LLMs. Furthermore, we introduce a dynamic pruning strategy incorporating an improved upper confidence bound~\citep{Srinivas2009InformationTheoreticRB} (UCB) formula to optimize the exploration-exploitation balance essential for effective decision-making in high-stakes tasks.

Our primary contributions are as follows:

We develop and validate a novel reasoning algorithm by integrating LLMs with UCT-MCTS. We enhance the algorithm's key components to accommodate the integration with LLMs better and demonstrate its effectiveness on Olympic-level mathematical problems.

We propose a dynamic pruning module that refines decision-making processes within the MCTS framework, facilitating more efficient and accurate problem-solving capabilities.

Through extensive experimentation, we provide insights into the synergistic potential of LLMs and MCTS, showcasing improved performance in complex reasoning tasks.

This research advances the application of LLMs in sophisticated reasoning challenges. It sets the stage for future innovations in integrating AI technologies for enhanced decision-making, reasoning accuracy, and reliability of LLM-driven applications.

\section{Preliminary}

This section introduces the preliminary and notations used in this work. We will first detail the mechanism of Monte Carlo Tree Search (MCTS) and, essential for understanding the novel dynamic Monte Carlo Tree Self-refine 


\textbf{Monte Carlo Tree Search (MCTS)} is a decision-making algorithm widely used in games and complex decision processes, which operates by building a search tree and simulating outcomes to estimate the value of actions. It involves four key phases~\citep{Browne2012ASO}: Selection, based on the UCT strategy to maximize the potential; expansion, where new nodes are added; simulation, to foresee possible outcomes; and Backpropagation, updating the node values based on simulation results. Typically, the MCTS algorithm comprises four distinct phases:
\begin{itemize}
\item \textbf{Selection:} Starting from the root, the algorithm navigates through promising child nodes based on specific strategies (e.g., UCT), continuing until a leaf node is reached. 

\item \textbf{Expansion:} At the leaf node, unless it represents a terminal state of the game, one or more feasible new child nodes are added to illustrate potential future moves.

\item \textbf{Simulation or Evaluation:} From the newly added node, the algorithm conducts random simulations—often termed "rollouts"—by selecting moves arbitrarily until a game's conclusion is reached, thereby evaluating the node's potential.

\item \textbf{Backpropagation:} Post-simulation, the outcome (win, loss, or draw) is propagated back to the root, updating the statistical data (e.g., wins, losses) of each traversed node to inform future decisions.
\end{itemize}
Repeatedly iterating through these stages, MCTS incrementally constructs a decision tree, refining strategies for optimal decision-making in scenarios where direct calculation of the best strategy is infeasible due to the vastness of the state space.

\textbf{Upper Confidence Bound applied on Trees} Algorithm is crucial for the selection phase in MCTS, balancing exploration and exploitation by choosing actions that maximize: 

\begin{equation}
 UCT_j = \bar{X}_j + C  \sqrt{\frac{2 \ln N_C}{N_j}} 
 \label{eq:1}
\end{equation}

Where \( \bar{X}_j \) is the average reward of action \( j \), \( N_C \) is the total visited times of the father node, and \( n_j \) is the number of times node \( j \) has been visited for simulation, C is a constant to balancing exploitation and exploration.

\textbf{MCT Self-Refine} algorithm represents an integration of Monte Carlo Tree Search (MCTS) with large language models, abstracting the iterative refinement process of mathematical problem solutions into a search tree structure. Nodes on this tree represent different versions of answers, while edges denote attempts at improvement. This algorithm's operational workflow adheres to the MCTS algorithm's general pattern. Detailly, We employ self-reflective driven self-improvement for refining answers; rewards for different answer versions are sampled using the model's self-reward capability. 


To facilitate understanding of the MCTSr algorithm, the following symbols and functions are defined:
\begin{itemize}
\item \( P \): The problem instance being addressed.

\item \( A \): The set of nodes, each representing a potential answer to \( P \).
\item \( M \): The set of actions available at each node, representing possible self-refine modifications to an answer.

\item \( R \): A function that samples self-rewards for nodes based on the quality and effectiveness of the modifications.

\item \( R_a \): A Set that stores all self-rewards sampling results of node $a$ with self-rewards function $R$.

\item  \( T \): A function determining the termination of the search process based on criteria such as reaching a maximum number of iterations or achieving satisfactory answer quality.

\item \( Q(a) \): A value function estimating the worth of an answer node \( a \), derived from accumulated rewards $R_a$ and backpropagations from children nodes.

\item \( U(a) \): The Upper Confidence Bound for the $Q$ value of node \( a \) to balance between exploitation and exploration.

\item \( \text{Father}(a) \): A function returning the parent node of a given node \( a \). If \( a \) is a root node, this function returns null or a specific identifier.

\item \( \text{Children}(a) \): A function returning the set of all child nodes for a given node \( a \), representing all possible states derived from \( a \) by executing actions \( m \in M \).

\item \( N(a) \): The total number of visits to node \( a \), used to calculate its UCB value and assess exploration and exploitation status. Since we will sample a reward for each visit, this value equals $|R_a|$. 
\end{itemize}

\section{Methodology}
In this section, we will first demonstrate the main structure of MCTSr, shown in Figure \ref{fig:main pic}. Then, we will detail each component
The main workflow of MCTSr is structured as follows:
\begin{itemize}
\item \textbf{Initialization}: A root node is established using either a naive model-generated answer and a dummy response (e.g., 'I don't know.') to minimize model overfitting tendencies.

\item \textbf{Selection}: The algorithm employs a value function $Q$ to rank all answers that were not fully expanded and selects the highest-valued node for further exploration and refinement using a greedy strategy.

\item \textbf{Self-Refine}: The selected answer $a$ undergoes optimization using the Self-Refine framework~\citep{Madaan2023SelfRefineIR}. Initially, the model generates a feedback $m$, guiding the refining process to produce an enhanced answer $a^{\prime}$.

\item \textbf{Self-Evaluation}: The refined answer is scored to sample a reward value and compute its $Q$ value. This involves model self-reward feedback and constraints such as strict scoring standards and suppression of perfect scores to ensure reliability and fairness in scoring.

\item \textbf{Backpropagation}: The value of the refined answer is propagated backward to its parent node and other related nodes to update the tree's value information. If the $Q$ value of any child node changes, the parent node's $Q$ is updated.

\item \textbf{UCT update}: After the $Q$ values of all nodes are updated, we identify a collection $\mathbf{C}$ of candidate nodes for further expansion or Selection, then use the $UCT$ update formula to update the $UCT$ values of all nodes for the next \textbf{Selection} stage.
\end{itemize}

The algorithm iterates through these stages until a termination condition 
 $T$ is met, including rollout constraints or maximum exploration depth, continuously refining the quality of answers, and exploring new possibilities.


 \subsection{Self-Refine}
In the self-refine process, the model is guided by a multi-turn dialogue refine prompt to optimize an answer \(a\) to problem \(P\). Initially, the model generates a reflective or critical comment \(m\) regarding \(a\). Subsequently, guided by \(m\), the model modifies \(a\) to produce an improved version \(a'\). This iterative refinement enhances the quality of the response, leveraging structured feedback to drive the evolution of the answer.


\subsection{Self-Evaluation}

In the refining process for mathematical problem \( P \), the \( Q \) value of an answer \( a \) is defined as the expected quality of further refining \( a \) into a superior answer, owing to the Markovian nature of the transition from \( a \) to its rewritten forms. Unlike traditional MCTS where \( Q(s, a) \) estimates the value of action \( a \) in state \( s \), \( Q(a) \) here derives from multiple samplings of the reward function values attributed to \( a \).

The model utilizes a self-reward method to estimate rewards for \( a \), where it is required to provide a reward score ranging from -100 to 100. We find that without constraints, the model's reward tendency is overly smooth, leading to a lack of comparative distinction between answers in practice. To address this, three constraints are designed:

\begin{itemize}

\item \textbf{Prompt Constraint}: The model must adhere to the strictest standards during reward scoring.
\item \textbf{Full Score Suppression}: The model is instructed not to provide full feedback scores; any reward above 95 is reduced by a constant amount to curb excessive scores.
\item \textbf{Repeated Sampling}: Each visit to a search tree node involves the repeated sampling of the node's rewards to enhance the reliability of the Self-Evaluation. It should be noted that when reward sampling is performed on the child nodes of a node, we will also perform reward sampling on its parent node to increase the sample size of reward sampling.
\end{itemize}

Post sampling, the \( Q \) value of \( a \) is calculated. To counteract the smoothing tendency of the self-reward function, a minimum value constraint is added to the expected reward, further refining the estimation of answer quality,

\begin{equation}
Q(a) = \frac{1}{2}\left(\min{R_a} + \frac{1}{|R_a|}\Sigma^{|R_a|}_{i=1}{R_a^i}\right)
 \label{eq:2}
\end{equation}

where \(Q(a)\) is the quality value of answer \(a\), \(R_a\) is the set of reward samples for \(a\), \(\min{R_a}\) is the minimum reward in \(R_a\), \(|R_a|\) is the number of samples, and \(\Sigma^{|R_a|}_{i=1}{R_a^i}\) is the sum of all rewards in \(R_a\). This formula calculates \(Q(a)\) by averaging the rewards' minimum and mean, balancing worst-case and average outcomes.
\subsection{Backpropagation}

After all leaf nodes' reward value sampling and Q value update are completed, we will propagate this change to its parent and ancestor nodes. During this update process, if the $Q$ function value of any element in the child node set $\text{Children}(a)$ of a node $a$ changes, the $Q$ function value of the node is updated to

\begin{equation}
Q^{\prime}(a) = \frac{1}{2} \left(Q(a) + \max_{i \in \text{Children}(a)} Q(i)\right)
 \label{eq:3}
\end{equation}

Where \(Q^{\prime}(a)\) is the updated quality value of answer \(a\) that consider the impact from its children nodes, \(Q(a)\) is the naive quality value only consider its reward samplings, and \(\max_{i \in \text{Children}(a)} Q(i)\) represents the highest quality value among the children of \(a\). This formula refines \(Q(a)\) by averaging the current value and the best possible outcome from its subsequent Children nodes.


\subsection{Update UCT and Selection}
After updating the \(Q\) values across all nodes in the tree, we proceed to the selection phase for the next round of choices. This process includes the following steps:

\textbf{Candidate Node Selection:} Leveraging the Markovian nature of the mathematical problem refine process, we focus on selecting all leaf nodes and those that are not fully expanded, disregarding the history of refine paths is feasible. This path-independent property helps simplify our problem. We no longer need to start from the root node when selecting nodes but traverse the nodes in the tree in hierarchical order.

But given that Large Language Models (LLMs), which play as policy in this task, can generate an infinite number of refine actions \(m\) for any answer state \(a\), each node potentially faces an unbounded set of actions for expansion. Thus, drawing from the concept of Expectation Improvement in Bayesian optimization, we propose two criteria for determining "full expansion":
\begin{itemize}
    \item The node's children count reaches a predefined limit. And,
   \item At least one child node’s \(Q\) value exceeds the node's.
\end{itemize}

We identify a collection $\mathbf{C}$ of candidate nodes based on these criteria for further expansion or Selection. This strategy helps accurately define which nodes might yield higher-value answers in subsequent searches, enhancing overall search efficiency and outcome quality.

\textbf{UCT Update:} 
Drawing from AlphaGo, we use UCT with the UCB-1 method to balance the exploration and exploitation of nodes; for node $a$ in the candidate set $\mathbf{C}$, its $UCT_a$ value is,
\begin{equation}
UCT_a = Q(a) + c \sqrt{\frac{\ln{N(\text{Father}(a)) + 1}}{N(a) + \epsilon}}
 \label{eq:4}
\end{equation}

where $Q(a)$ is the $Q$ value of answer $a$, $N(\cdot)$ is the total visited times of given nodes, $c$ is a constant to balancing exploitation and exploration, $\epsilon$ is a small constant for avoid devided-by-zero.

\textbf{Sorting and Selection:} 
According to the UCT value of the candidate set $\mathbf{C}$, we can select an optimal node to explore the refining process through greedy sampling or importance sampling.

\subsection{Termination Function}
In MCTSr algorithms, search termination function criteria $T$ can derive from several conditions:

\textbf{Early Stopping:} Termination occurs when improvements in search results diminish or when consecutive searches yield repetitive outcomes.

\textbf{Search Constraints:} The search terminates once the number of rollouts reaches a predetermined limit or when one or more nodes in the tree satisfy the maximum depth constraint.

\textbf{Advanced Criteria Based on Language Model Logits:} The search concludes based on predefined metrics derived from the language model's logits.

Once the Termination Function condition $T$ is satisfied, we can gather the best answers from tree nodes according to $Q$ values or other conditions.

\section{Evaluation}
\subsection{Experiment Settings}
To assess the MCTSr algorithm's effectiveness in solving mathematical problems, we employed LLaMA3-8B~\citep{llama38b} as the foundational model, enhanced with MCTSr. Detailed prompt settings are provided in the appendix. We compared LLaMA3-8 B's performance across several configurations—Zero-Shot CoT~\citep{Wei2022ChainOT}, Self-Refine, 4-rollouts MCTSr, and 8-rollouts MCTSr—against the performances of GPT-4~\citep{achiam2023gpt}, Claude 3~\citep{anthropic2024claude} and Gemini 1.5-Pro~\citep{reid2024gemini}, which are the latest state-of-the-art closed-source models. These comparisons were conducted on various datasets, including GSM8K~\citep{cobbe2021training}, GSM Hard~\citep{gao2022pal}, MATH~\citep{DBLP:journals/corr/abs-2103-03874}, AIME~\citep{noauthor_aime_nodate}, Math Odyssey~\citep{noauthor_agi_nodate}, and OlympiadBench(pure-text subset)~\citep{he2024olympiadbench}.

\subsection{GSM Benchmarks}
We evaluated the above methods on the test sets of GSM8K and GSM-hard, which involved typical and challenging mathematical problems, respectively. The results are shown in Table~\ref{tab: gsm result}. 

We can find that results reveal a direct correlation between the number of MCTSr rollouts and success rates, significantly improving as iterations increase, especially in the less complex GSM8K. However, the more intricate GSM-Hard set showcased a performance ceiling even at higher rollouts, indicating the limits of current strategies against complex problems.

These insights underscore the MCT-Self-refine algorithm's robustness and potential boundaries, highlighting the necessity for ongoing enhancements to tackle more complex challenges effectively. This work demonstrates the algorithm's capacity to enhance problem-solving performance and its varying efficacy across problem complexities, suggesting areas for future refinement in educational technology and automated reasoning.

\begin{table}[]
\centering
\begin{tabular}{c|c|c|c|c|c}
\hline
\hline
\multirow{2}{*}{Datasets} & Zero-Shot & One-turn    & 4-rollouts & 8-rollouts & Example                \\
                          & CoT       & Self-refine & MCTSr      & MCTSr      & Nums                   \\ \hline
\multirow{2}{*}{GSM8K}    & 977       & 1147        & 1227       & 1275       & \multirow{2}{*}{1319}  \\ \cline{2-5}
                          & 74.07\%   & 86.96\%     & 93.03\%    & 96.66\%    &                        \\ \hline
\multirow{2}{*}{GSM-Hard} & 336       & 440         & 526        & 600        & \multirow{2}{*}{1319}  \\ \cline{2-5}
                          & 25.47\%   & 33.36\%     & 39.88\%    & 45.49\%    &                        \\ \hline \hline
\end{tabular}
\newline\caption{Performance of MCTSr on the GSM Dataset}
\label{tab: gsm result}
\end{table}

\subsection{MATH Benchamark}

This section presents the outcomes of applying the MCT-Self-refine (MCTSr) algorithm across various complexity levels on the MATH dataset. The dataset is stratified into five levels of difficulty, ranging from level 1 (easiest) to level 5 (most challenging). The algorithm's performance is evaluated using four distinct configurations: Zero-Shot CoT, One-turn Self-refine, 4-rollouts MCTSr, and 8-rollouts MCTSr. Each configuration's efficacy is measured by the number of successfully solved problems and the corresponding success rates, with a total of 5000 examples across all levels.

Level-1 results demonstrate the highest success rates, with the 8-rollouts MCTSr achieving a remarkable 90.16\% success rate, solving 394 out of 437 problems. This level shows a clear progression in success rates as rollouts increase.

At the most challenging level-5 part, the 8-rollouts MCTSr configuration yields a 34.06\% success rate, solving 451 out of 1324 problems. This illustrates the increasing difficulty and the algorithm's strained performance in highly complex scenarios.

Overall performance across all levels shows a cumulative success rate of 58.24\% with the 8-rollouts MCTSr, solving 2912 out of 5000 problems. This rate demonstrates a substantial enhancement from the Zero-Shot CoT's initial rate of 24.36\%. The data indicates a consistent trend where the increase in rollouts correlates with improved success rates, underlining the efficacy of the MCT-Self-refine algorithm in enhancing problem-solving capabilities across varying levels of mathematical complexity.

These results validate the MCT-Self-refine algorithm's potential in academic and problem-solving contexts and highlight its scalability and adaptability to different levels of problem complexity within the MATH dataset.

\begin{table}[]
\centering
\begin{tabular}{c|c|c|c|c|c}
\hline\hline
\multirow{2}{*}{Level}   & Zero-Shot & One-turn    & 4-rollouts & 8-rollouts & Example               \\
                         & CoT       & Self-refine & MCTSr      & MCTSr      & Nums                  \\ \hline
level-1                  & 250       & 314         & 365        & 394        & \multirow{2}{*}{437}\\ \cline{2-5}
                         & 57.21\%   & 71.85\%     & 83.52\%    & 90.16\%    &                       \\ \hline
\multirow{2}{*}{level-2} & 363       & 474         & 594        & 692        & \multirow{2}{*}{894}  \\ \cline{2-5}
                         & 40.60\%   & 53.02\%     & 66.44\%    & 77.40\%    &                       \\ \hline
\multirow{2}{*}{level-3} & 309       & 454         & 585        & 719        & \multirow{2}{*}{1131} \\ \cline{2-5}
                         & 27.32\%   & 40.14\%     & 51.72\%    & 63.57\%    &                       \\ \hline
\multirow{2}{*}{level-4} & 202       & 368         & 523        & 656        & \multirow{2}{*}{1214} \\ \cline{2-5}
                         & 16.64\%   & 30.31\%     & 43.08\%    & 54.04\%    &                       \\ \hline
\multirow{2}{*}{level-5} & 94        & 177         & 290        & 451        & \multirow{2}{*}{1324} \\ \cline{2-5}
                         & 7.10\%    & 13.37\%     & 21.90\%    & 34.06\%    &                       \\ \hline
\multirow{2}{*}{Overall} & 1218      & 1787        & 2357       & 2912       & \multirow{2}{*}{5000} \\ \cline{2-5}
                         & 24.36\%   & 35.74\%     & 47.14\%    & 58.24\%    &                       \\ \hline\hline
\end{tabular}
\newline\caption{Performance of MCTSr on the MATH Dataset}
\label{tab:MATH result}
\end{table}
\subsection{Olympiad-level Benchmarks}

The efficacy of the MCT-Self-refine (MCTSr) algorithm was tested on three datasets from mathematical Olympiad competitions: AIME, GAIC Math Odyssey, and OlympiadBench. The GAIC Math Odyssey dataset, released in April 2024, is notable for its minimal overlap with the pre-training corpus of the LLaMa3-8B model, providing a robust test of the algorithm's ability to generalize.

\textbf{AIME:} From Zero-Shot CoT's 2.36\% (22 problems solved) to 8-rollouts MCTSr's 11.79\% (110 problems solved).

\textbf{GAIC Math Odyssey:} Showed substantial improvement, starting at 17.22\% (67 problems solved) and reaching up to 49.36\% (192 problems solved) with 8-rollouts MCTSr.

\textbf{OlympiadBench:} Improved from 1.25\% (16 problems solved) in Zero-Shot CoT to 7.76\% (99 problems solved) in 8-rollouts MCTSr.

The results demonstrate a clear trend where increased rollouts correlate with higher success rates, highlighting the algorithm's potential to improve performance through iterative refinement. The GAIC Math Odyssey results mainly reflect the MCTSr's generalization capabilities in new environments.

These findings affirm the MCT-Self-refine algorithm's robustness and its utility in tackling complex, unseen mathematical problems, suggesting its applicability in educational technologies aimed at competitive academic settings like Olympiads.

\begin{table}[]
\centering
\begin{tabular}{c|c|c|c|c|c}
\hline\hline
Datasets &
  \begin{tabular}[c]{@{}c@{}}Zero-Shot \\ CoT\end{tabular} &
  \begin{tabular}[c]{@{}c@{}}One-turn \\ Self-refine\end{tabular} &
  \begin{tabular}[c]{@{}c@{}}4-rollouts \\ MCTSr\end{tabular} &
  \begin{tabular}[c]{@{}c@{}}8-rollouts \\ MCTSr\end{tabular} &
  \begin{tabular}[c]{@{}c@{}}Example \\ Nums\end{tabular} \\ \hline
\multirow{2}{*}{AIME}          & 22      & 41      & 70      & 110     & \multirow{2}{*}{933}  \\ \cline{2-5}
                               & 2.36\%  & 4.39\%  & 7.50\%  & 11.79\% &                       \\ \hline
\multirow{2}{*}{Math Odyssey}  & 67      & 118     & 156     & 192     & \multirow{2}{*}{389}  \\ \cline{2-5}
                               & 17.22\% & 30.33\% & 40.10\% & 49.36\% &                       \\ \hline
\multirow{2}{*}{OlympiadBench} & 16      & 39      & 67      & 99      & \multirow{2}{*}{1275} \\ \cline{2-5}
                               & 1.25\%  & 3.06\%  & 5.25\%  & 7.76\%  &                       \\ \hline\hline
\end{tabular}
\newline\caption{Performance of MCTSr on Olympiad-level Datasets}
\label{tab: olympiad result}
\end{table}

\subsection{Disscussion}
We investigated the reported values of the current state-of-the-art closed-source large model SOTA performance on the above test benchmarks, as shown in the table~\ref{tab: closed-source LLM}.
\begin{table}[]
\centering
\begin{tabular}{c|c|c|c}
\hline\hline
 &
  \begin{tabular}[c]{@{}c@{}}Gemini\\ 1.5-Pro\end{tabular} &
  \begin{tabular}[c]{@{}c@{}}Claude \\ 3 Opus\end{tabular} &
  \begin{tabular}[c]{@{}c@{}}GPT-4\\ Turbo\end{tabular} \\ \hline
MATH~\citep{reid2024gemini}              & 67.7 & 60.1 & 73.4 \\ \hline
Math Odyssey~\citep{reid2024gemini} & 45.0 & 40.  & 49.1 \\ \hline
GSM8K~\citep{noauthor_papers_nodate}             & 94.4 & 95   & 97.1 \\ \hline\hline
\end{tabular}
\caption{closed-source LLM performance on mathematical datasets}
\label{tab: closed-source LLM}
\end{table}
Comparing the performance of current closed-source large models, MCTSr can effectively enhance the mathematical reasoning capabilities of small-parameter open-source models, like LLaMa-3, to a comparable level.
\section{Related works}

Monte Carlo Tree Search (MCTS) has been widely utilized in various fields to solve complex problems efficiently. \cite{pitanov2023monte} explored the application of MCTS in Multi-agent Pathfinding, demonstrating its superiority over heuristic search algorithms like A*. Additionally, \cite{yang2023integrated} integrated MCTS with heuristic, unsupervised, and supervised learning methods to efficiently solve the Train Timetabling Problem (TTP). Furthermore, \cite{li2023general} introduced a general method for solving various types of SAT problems using a unified framework incorporating MCTS. \cite{vagadia2024phyplan} developed PhyPlan, a physics-informed planning framework that combines physics-informed neural networks with modified MCTS to enable robots to perform dynamic physical tasks effectively. In conclusion, MCTS has proven to be a versatile and effective mathematical solution for solving various complex problems in different domains, including robotics, game solving, and optimization. Researchers continue to explore and enhance the capabilities of MCTS by integrating it with other algorithms and frameworks to tackle increasingly challenging tasks.

Recent research has made notable strides in enhancing mathematical reasoning in large language models (LLMs). \cite{du2023improving} introduced a method where multiple LLMs discuss and refine answers collectively, significantly boosting reasoning and factual accuracy. \cite{luo2023wizardmath} developed WizardMath, which leverages Reinforcement Learning from Evol-Instruct Feedback to surpass existing LLMs in mathematical benchmarks. Meanwhile, \cite{lu2023mathvista} created MathVista, a visual, mathematical benchmark, with GPT-4V achieving a 49.9\% accuracy, highlighting gaps that persist relative to human performance. \cite{yu2023metamath} introduced MetaMath, a fine-tuned model that excels in mathematical challenges, and \cite{yuan2023scaling} demonstrated that pre-training loss and Rejection sampling Fine-Tuning can optimize LLM performance, particularly in less advanced models. These studies suggest significant progress yet underline the necessity for ongoing research in LLM mathematical reasoning.

Recent advancements in large language models (LLMs) have significantly improved their mathematical reasoning abilities. Yet, they still face complex problems that require multiple reasoning steps, leading to logical or numerical errors. To address this limitation, \cite{chen2024alphamath} proposed incorporating Monte Carlo Tree Search (MCTS) to enhance the mathematical reasoning capabilities of fine-tuned LLMs without additional fine-tuning steps . \cite{xu2023no} utilized MCTS and a lightweight energy function, the models can rank decision steps and enable immediate reaction and precise reasoning, leading to improved performance on mathematical reasoning benchmarks. However, it still lacks a framework that combines the self-refine capabilities and self-reward evaluation method of LLMs to refine the model's response iteratively with the Monte Carlo Tree Search Algorithm.

\section{Limitations}
Although the MCTSr algorithm has demonstrated certain advantages in mathematical tasks, our research is still in its preliminary stages. As a general decision-making framework, the potential applications of MCTSr in various scenarios remain to be explored further, such as in black-box optimization problems and self-driven alignment for large language models. Additionally, the components of MCTSr are highly scalable, necessitating ongoing development to identify and compare a broader range of component algorithms, thereby enhancing the practical potential and effectiveness of the MCTSr algorithm.

\section{Conclusion}
This paper demonstrates the effectiveness of the MCT Self-Refine (MCTSr) algorithm in enhancing the capability of Large Language Models (LLMs) to solve complex mathematical problems. By integrating Monte Carlo Tree Search (MCTS) with LLMs, MCTSr addresses critical challenges in accuracy and reliability, particularly within mathematical reasoning tasks. Experimental results confirm significant improvements in problem-solving success rates across multiple datasets, including notable performance in Olympic-level mathematical challenges.

Moreover, the research advances the application of LLMs in sophisticated reasoning tasks and lays the groundwork for future integration of AI technologies to enhance decision-making and reasoning accuracy. Despite MCMCTSr'semonstrated potential in mathematical problem-solving, its applicability in broader contexts, such as black-box optimization and self-driven alignment, remains to be explored. Future work will optimize algorithmic components and test their performance across various problems and settings to achieve broader practicality and effectiveness.
\bibliographystyle{apalike}
\bibliography{reference}

\appendix

\section{Prompts in Experiment}
\subsection{Self-Refine}
\textbf{Get Feedback:}
\xiaoshuicomment{Since we have a weak Answer, could you provide me with a relection or feedback to correct this answer better? Analyze this Answer Strictly and Critic, point out every flaw for ervery possible imperfect to minus every possible score!

Let's think step by step.}

\textbf{Get Refined Answer:}
\xiaoshuicomment{
Please refine the your answer according to your Reflection or Feedback. The response should begin with [reasoning process]...[Verification]... and end with end with "[Final Answer] The answer is [answer formula]"

Let's think step by step.
}

\subsection{Self-Reward}
\xiaoshuicomment{Question: {question}

Answer: {ans}

Analyze this Answer Strictly and Critic, and point out every flaw for every possible imperfect to minus every possible score! You need to be very harsh and mean in calculating grades, and never give full marks to ensure that the marks are authoritative. 

Output a score between [-100,+100], ig. from -100 to +100. 

Response format:

[Analyst]...[Score]...}
\subsection{Dummy Answers}
\xiaoshuicomment{["I Don't Know","I can't understand this question.","I can't help with this question.","I don't know how to solve this question.","I don't know the answer to this question.","I don't know the answer to this question, sorry."]}

\end{document}